\documentclass{article}

\usepackage{booktabs}
\usepackage[table,xcdraw]{xcolor}
\usepackage[final, nonatbib]{neurips_2022}

\usepackage[utf8]{inputenc} 
\usepackage[T1]{fontenc}    
\usepackage{hyperref}       
\usepackage{url}            
\usepackage{booktabs}       
\usepackage{amsfonts}       
\usepackage{nicefrac}       
\usepackage{microtype}      
\usepackage{xcolor}         

\usepackage{xcolor} 
\usepackage{biblatex}

\newcommand{\todo}[1]{}
\renewcommand{\todo}[1]{{\textcolor{magenta}{TODO: {#1}}}}
\newcommand{\inlinecomment}[1]{}
\renewcommand{\inlinecomment}[1]{{\textcolor{blue}{COMMENT: {#1}}}}

\title{The Design Space of Generative Models}

\author{
  Meredith Ringel Morris \\
  Google Research\\
  Seattle, WA, USA \\
  \texttt{merrie@google.com} \\
  \And
  Carrie J. Cai \\
  Google Research \\
  Mountain View, CA, USA \\
  \texttt{cjcai@google.com} \\
  \And
  Jess Holbrook\thanks{work done while employed at Google Research} \\
  Meta \\
  Seattle, WA, USA \\
  \texttt{jessholbrook@fb.com} \\
  \And
  Chinmay Kulkarni \\
  Google Research \\
  Atlanta, GA, USA \\
  \texttt{ckulkarni@google.com} \\
  \And
  Michael Terry \\
  Google Research \\
  Cambridge, MA, USA \\
  \texttt{michaelterry@google.com} \\
}

\begin{document}

\maketitle

\begin{abstract}
  Card \textit{et al.'s} classic paper "The Design Space of Input Devices" \cite{card_design_space} established the value of design spaces as a tool for HCI analysis and invention. We posit that developing design spaces for emerging pre-trained, generative AI models is necessary for supporting their integration into human-centered systems and practices. We explore what it means to develop an AI model design space by proposing two design spaces relating to generative AI models: the first considers how HCI can impact generative models (i.e., interfaces for models) and the second considers how generative models can impact HCI (i.e., models as an HCI prototyping material).   
\end{abstract}

\section{Introduction}

Among the most significant advances in machine learning in recent years are the introduction of \textit{pre-trained, generative AI models} (sometimes called "foundation models" \cite{foundation_models}); these general-purpose models are trained on large data sets and can generate novel text (e.g., \cite{gpt3_paper}, \cite{lamda_arxiv}), imagery (e.g., \cite{dalle2}, \cite{imagen}), or other media that is in many cases indistinguishable in quality from human-created content \cite{clark2021}. Because of the relative ease with which they can be customized and controlled (e.g., through text-based ``prompts''), these models offer a number of potential benefits to human-centered computing, particularly for the field of Human-Computer Interaction (HCI). More specifically, generative models have the potential to revolutionize design and evaluation methodologies for a wide variety of interactive systems, as well as to support novel interaction paradigms (e.g., generative language models' potential to support fast, accurate communication for people with mobility \cite{naacl_abbrev_paper} and/or cognitive \cite{lampost} disabilities). 

In this position paper, we propose two preliminary ``design spaces'' \cite{card_design_space}: one for conceptualizing how HCI may impact the burgeoning field of generative model research, and a second for theorizing how generative models may impact HCI practices. We do not argue that these proposed design spaces are definitive or complete; rather, we hope that these serve as useful artifacts for reflection, discussion, and debate during the NeurIps 2022 Workshop on Human-Centered AI.

\section{Design Space for Interfaces to Generative Models}

Our first design space considers the application of HCI to generative AI models. The interfaces designed for interacting with such models have the potential to influence who is able to use the models (e.g., ML engineers vs. general users), model safety (e.g., supporting identification or interrogation of undesired outputs), and model applications (e.g., by altering the time and effort involved in prompt engineering). Our proposed taxonomy (Table~\ref{tab:llm-design-space}) considers both the design possibilities for interfaces for providing input to models and those for presenting model outputs.

\begin{table}[htbp]
\begin{tabular}{@{}p{2cm}|p{2cm}p{2cm}p{2cm}p{2cm}p{2cm}@{}}
\toprule
\multicolumn{6}{c}{\textbf{Input Dimensions}}\\ \midrule \\
Media & language & image & video & audio & multimodal \\
\hline
Format & free-form & restricted choice & templated & edits \\
\hline
Explainability & \begin{tabular}[c]{@{}l@{}} input \\ tokenization\end{tabular} & \begin{tabular}[c]{@{}l@{}} input \\ frequency\end{tabular} & blocklists \\
\hline
Evolution & prompting & prompt-tuning & fine-tuning & retraining \\
\bottomrule 
\toprule
\multicolumn{6}{c}{\textbf{Output Dimensions}} \\
\midrule \\
Media & language & image & video & audio & multimodal \\
\hline
Product & artifact (\textit{text}, \textit{image}, etc.) & customized model & prompt(s) \\
\hline
Timing & asynchronous & real-time \\
\hline
Operation & classification & completion & creation & information extraction & transformation \\
\hline
Interactions & \begin{tabular}[c]{@{}l@{}} quality control \\(\textit{bug reporting}, \\ \textit{safety reporting}) \end{tabular} & input iteration & query \\
\hline
Explainability & link output to input & link output to training data & \begin{tabular}[c]{@{}l@{}} metadata \\ embedding\end{tabular} & transparency (\textit{constrained decoding}, \textit{safety filtering})\\
\hline
Format & \begin{tabular}[c]{@{}l@{}} detail  (\textit{safety} \\ \textit{warnings},\\ \textit{confidence} \\ \textit{values}, etc.) \end{tabular} & quantity (\textit{top-1}, \textit{top-n}) & stylistic choices (e.g., \textit{anthropomorphization}) & standardization \\

\bottomrule 
\end{tabular} \\
\caption{Our proposed design space for interfaces to generative models; the first section enumerates the range of design parameters for input interfaces to models, while the second section demonstrates the range of design possibilities for presenting model outputs.}
\label{tab:llm-design-space}
\end{table}

\subsection{Design Space for Model Inputs}
When considering model inputs, different \textit{media} are possible; note that the input media and the output media may not necessarily match (e.g., the use of language prompts to generate images in status quo systems like DALL-E \cite{dalle2} and Imagen \cite{imagen}, though as our design space suggests, one can imagine alternative media input to output pairings such as submitting a sketch or musical score as input to generate an image). 

Varying input \textit{formats} may also be suitable for different purposes -- while many of today's language models (e.g., \cite{gpt3_paper}, \cite{lamda_arxiv}) allow free-form input in the form of an open-ended textbox, alternatives might include restricting choices via a menu (e.g., similar to many customer service chatbots or as done in the original Dall-E announcement blog\footnote{https://openai.com/blog/dall-e/}) or offering a template to help make the many potential parameters of complex prompts more apparent to novice users. Editing an existing artifact or collection of artifacts (including those previously output by the system) may be another way to specify model inputs (e.g., circling the portions of a generated image that did not meet the user's initial expectations). 


While status quo interfaces to models offer little or no feedback to end-users, future tools might provide \textit{explainability} features for model inputs, such as providing the end-user with information about input tokenization, frequency, block-listed items, etc., which might grow the user's mental model of the system and support improved utilization. 

Finally, our taxonomy accounts for the ways in which the input might support the \textit{evolution} of the model itself, whether via prompting, prompt-tuning, fine-tuning, or retraining.

\subsection{Design Space for Model Outputs}
When considering model outputs, in addition to generating artifacts comprising one or more \textit{media}, our design space also considers other kinds of \textit{products} that may result from the user's interactions; for instance, while GPT-3 might produce a paragraph of text as an artifact, end-users may be equally or more interested in the customized model created via a given prompting strategy, or in the prompt(s) used to produce a given artifact. 

\textit{Timing} of outputs may also be varied; while real-time results are necessary for supporting many interactions and experiences, asynchrony may be necessary for computationally expensive models (e.g., video generation), yet may offer new design opportunities (e.g., previewing or giving feedback on partial outputs). 

Interfaces for model output may also account for the \textit{operation} performed by the model; common operations include classification (e.g., a language model that can tell you if an input word is or is not a type of food), completion (e.g., a model that can provide a punchline given the setup for a joke), creation (e.g., producing a short story), information extraction (e.g., a question-answering system that allows the user to probe the underlying training data), and transformation (e.g., a model that turns English text into French). Additional fundamental operations are likely to be discovered as new use cases and model capabilities evolve. 

Today's interfaces to model outputs tend to be static, but we envision a future where interfaces afford \textit{interactions} with outputs, including allowing the end-user to give quality control feedback (e.g., report a bug or safety issue), to iterate by refining or editing the output and resubmitting it to the  model for further processing, or to query the output in order to understand more about why a particular output or portion of an output was generated. 

Interactive querying is one possible instantiation of \textit{explainability} - the presence or absence of explainability features (and their manner of presentation to the end-user) should be a key dimension of model output interfaces, including mechanisms for explaining how outputs link to inputs and how they link to training data. Further explainaiblity features might include whether metadata (such as the model version, prompt, and any model parameters) are embedded in resultant artifacts, and whether the system offers transparency into any filtering stages that may impact output (e.g., constrained decoding, safety filtering, etc.). 

Finally, there are many options for outputs' \textit{format} to the end-user. This includes stylistic choices, such as whether to employ anthropomorphization (as is common in many chatbots). It also includes concerns around the level of detail to present to the end-user, such as whether and how to present safety warnings, confidence metrics, or other details related to the inner workings of the model. Additionally, the choice of the quantity of alternative outputs produced (e.g., is the user shown a single, canonical output or a set of \textit{n} to choose among) can greatly impact the user's experience and mental model regarding the determinism of the system. Finally, as the variety of available models continues to proliferate, it is important to consider \textit{standardization} of output formats, such that the output of one model might be consumed by other models (e.g., to support pipelining between models, and/or to other related tools). 

\subsection{Applying the HCI-for-Models Design Space}

As an example of how this taxonomy can be used to describe particular interfaces, consider PromptMaker \cite{macromaker}, an application intended to support prompt programming. Along the \textit{input} dimensions, PromptMaker accepts language (text) as input (media) in a free-form format, and provides explicit support for both zero- and few-shot prompting. It provides no explainability capabilities or evolution of the prompts (an oversight our taxonomy might make visible to the PromptMaker creaters, perhaps suggesting an area for future development!). Along the \textit{output} dimensions, PromptMaker produces language (text) as output (media), where this text is the final artifact (product). It is real-time (timing), and can support a number of types of operations (classification, completion, creation, information extraction). Built into the interface are safety reporting mechanisms (interactions) and safety warnings about the potential for undesirable content (presentation).

In addition to being descriptive, our taxonomy can also be generative, identifying new opportunities for design and research. For example, status quo model interfaces typically produce a static output, but \textit{interactive outputs}, such as those that support input iteration, could dramatically change the pace at which people are able to experiment with and create content using generative models. Altering the \textit{media} used as input could also afford novel experiences - sketch-based, photograph-based, or even audio inputs rather than natural-language strings could all afford new means of expression and creativity with generative image models. 

Many considerations, including the use context and target end-users, will factor into how to select among these design parameters when developing interfaces for generative models. While some dimensions may seem upon first glance to be inherently "good" or "bad," such choices are not always clear-cut. For instance, while there are many risks in inappropriate or undisclosed \textit{anthropomorphization} of AI, this may nevertheless be a desirable interface for some customer service or entertainment applications. On the other hand, while \textit{explainability} is generally viewed positively, consider the scenario of whether to include the \textit{output explainability} feature of metadata embedding in a generative image model. If the use context is in an application for professional artists creating original works, embedding the prompt as metadata may be undesirable, since the artist may view the prompt as their intellectual property. However, if the use context is in an application for generating custom clip-art for use in documents and slide decks, then embedding the prompt as metadata may be highly beneficial, since it could function as alternative text for screen readers, thereby ensuring that documents produced with this clip art are accessible to blind end-users.

\section{Design Space for Generative Models as an HCI Prototyping Material}

We believe that the increasing variety and power of generative models will fundamentally change HCI research and practice by enabling a new generation of assets, tools, and methods. These artifacts will impact the full spectrum of HCI, including new tools to support design through creative ideation \cite{wordcraft}, new tools to support building rapid prototypes of novel, interactive experiences \cite{lampost}, and new tools to support evaluation of interfaces and ecosystems, such as through simulation \cite{social_sim_uist}. Table~\ref{tab:hci-design-space} offers a starting point for such a design space.

\begin{table}[htb]
\begin{tabular}{@{}l|llll@{}}
\toprule
\textbf{Dimension }              & \multicolumn{3}{c}{\textbf{Design choices}}                                                                                                                                                                                                                                           &  \\ \midrule 
Goal      & \cellcolor[HTML]{C0C0C0}\begin{tabular}[c]{@{}l@{}}capture rich context \\ of use (“Field” \cite{zimmerman2014research})\end{tabular} &
\begin{tabular}[c]{@{}l@{}}co-create  with \\ users (“Studio” \cite{zimmerman2014research})\end{tabular} &
\begin{tabular}[c]{@{}l@{}} study specific\\  phenomena (“Lab” \cite{zimmerman2014research})\end{tabular} &   \\
\hline
Role of AI              & as-is capabilities & stand-in for future model                                                                                & \cellcolor[HTML]{C0C0C0}\begin{tabular}[c]{@{}l@{}}synthetic data (\textit{simulating input},\\ \textit{simulating content})\end{tabular}                                                                                                   &  \\
\hline
Context          & context-free                                                                                       & \cellcolor[HTML]{C0C0C0}{\color[HTML]{000000} minimal context}               & \begin{tabular}[c]{@{}l@{}} deep context \end{tabular}                                                         &  \\
\hline
Lifecycle               & \cellcolor[HTML]{C0C0C0}{\color[HTML]{000000}  inspiration}                                                   &  building                                                                      & evaluation                                                                         &  \\
\hline
Fidelity               & \cellcolor[HTML]{C0C0C0}{\color[HTML]{000000} proving }                                                   & scaling                                         & serving                                                                         &  \\
\hline
Media & \cellcolor[HTML]{C0C0C0}text                                                                             & visual media (\textit{images}, \textit{video})                                                                & sound                                                                              &  \\
 \bottomrule
\end{tabular}
\caption{Our proposed design space of generative models as a tool for  human-computer interaction practices. The shaded cells depict one particular choice, as in~\cite{social_sim_uist}, of a system that simulates other users to speed up prototyping of a social media platform interface.}
\label{tab:hci-design-space}
\end{table}

\subsection{Dimensions of the Models-for-HCI Design Space}

In this taxonmy, the \textit{goal} dimension acknowledges that generative models can be used for prototyping different kinds of design explorations. For instance, borrowing terminology from~\cite{zimmerman2014research}, generative models could allow designers to capture how a new AI-based system might behave in the rich context of the ``field,'' without building it out fully. Examples of prototypes with such a ``field'' goal include~\cite{social_sim_uist}, which uses an AI model to simulate the rich social context of the real world in which newsgroups and other feed interfaces are deployed. Instead of simulation, other systems focus on co-creation ("the studio" in \cite{zimmerman2014research}'s terminology); for instance, Wordcraft~\cite{wordcraft} uses a generative model to augment a story-writing interface. In this case, the goal of the prototype is to co-create fiction stories with the user (note how Wordcraft, unlike ``social simulacra''~\cite{social_sim_uist}, cannot fulfill its prototyping goal without real users' input). Finally, generative models may also help prototyping by narrowing the focus of investigation, or studying specific phenomena. One example of this "lab" goal is ``AI Chains''~\cite{wu2022ai}, which studied whether breaking up the operation of a language model into smaller steps and aggregating them allows for greater human control and transparency.

In using generative models as an HCI prototyping material, the \textit{role of AI} may also vary greatly. Models could be used for their currently known capabilities, as with many AI systems today using GPT-3. They could also represent a less developed version of a yet-to-be-created model, to investigate whether it is worth the effort of creating a specialized model.  Finally, they could allow researchers access to synthetic data, for instance by simulating user actions or populating a system with content produced by models.

The  \textit{context} dimension captures the idea that generative models can be used to prototype interactions with differing access to context. For instance, models may primarily perform tasks that are essentially  context-free (e.g., rewriting a text snippet to be more polite, generating images from text prompts, etc.). They may also work with minimal context (for instance, recommending a movie based on a list of movies the user likes or generating a proposed chat reply based on several turns of conversational history). Finally, generative models can use deep context accumulated over a long period of time, for instance, encoding rhetorical preferences in writing learned over months or years of interaction with a particular author. 

The \textit{lifecycle} dimension reflects the range of HCI practices that may be impacted by emerging models. For instance, models could support the early parts of the design process by producing content to inspire creativity (e.g., suggesting design scenarios); they could support the building stage of HCI systems by standing in for a more mature system component such as a fine-tuned or personalized model; or they could support the evaluation stage of HCI work by simulating user-generated content or interaction sequences. 

A related aspect is the \textit{fidelity} dimension, which suggests how emerging AI models can be suited to different aspects of HCI practice that require different levels of robustness, speed, safety, or other dimensions of performance. Appropriate fidelity levels may differ for scenarios such as to prove or disprove a design hypothesis, to scale an interaction to more people than possible through other prototyping methods (such as Wizard of Oz), or serving ``in production,'' for users to experience as a completed product. 

Finally, the \textit{media} (or combination of multiple media) supported by a model will, of course, shape its suitability and role as an HCI design material.

\subsection{Applying the Models-for-HCI Design Space}

As with our "interfaces for model inputs \& outputs" taxonomy, this second design space can also stimulate research, design, and reflection by supporting both descriptive and generative functions. The shading in (Table~\ref{tab:hci-design-space}) reflects a descriptive example, illustrating a classification for Park et al.'s "Social Simulacra" work \cite{social_sim_uist}, which used GPT-3 to populate parallel hypothetical futures of proposed social media systems in order to allow social media designers to tweak system features to facilitate desired interaction styles. 

This ontology can also suggest directions for future investigation of the potential for generative models to further the field of HCI. For instance, models that generate text could be used in the design stage of the HCI lifecycle to create personas that would inspire a practitioner to consider a wide range of design concepts. Similarly, language models could be used to co-create productivity tools with users, for instance, embedding the context of their work into writing aids in word processors. 

\section{Conclusion}

Card et al. \cite{card_design_space} established the value of design spaces and similar ontologies for analysis and invention. In this position paper, we seek to advance discussion and reflection around the marriage of HCI and emerging generative AI models by proposing two design spaces: one for HCI applied to generative models (Table~\ref{tab:llm-design-space}) and one for generative models applied to HCI (Table~\ref{tab:hci-design-space}). We hope these frameworks spark additional research, both in further refining these taxonomies as well as in spurring the development of novel interfaces for generative models and novel AI-powered design and evaluation tools for HCI broadly. 

\printbibliography

@incollection{zimmerman2014research,
  title={Research through design in HCI},
  author={Zimmerman, John and Forlizzi, Jodi},
  booktitle={Ways of Knowing in HCI},
  pages={167--189},
  year={2014},
  publisher={Springer}
}

@inproceedings{wu2022ai,
  title={Ai chains: Transparent and controllable human-ai interaction by chaining large language model prompts},
  author={Wu, Tongshuang and Terry, Michael and Cai, Carrie Jun},
  booktitle={CHI Conference on Human Factors in Computing Systems},
  pages={1--22},
  year={2022}
}

@inproceedings{card_design_space,
author = {Card, Stuart K. and Mackinlay, Jock D. and Robertson, George G.},
title = {The Design Space of Input Devices},
year = {1990},
isbn = {0201509326},
publisher = {Association for Computing Machinery},
address = {New York, NY, USA},
url = {https://doi.org/10.1145/97243.97263},
doi = {10.1145/97243.97263},
booktitle = {Proceedings of the SIGCHI Conference on Human Factors in Computing Systems},
pages = {117–124},
numpages = {8},
location = {Seattle, Washington, USA},
series = {CHI '90}
}

@misc{lamda_arxiv,
  doi = {10.48550/ARXIV.2201.08239},
  
  url = {https://arxiv.org/abs/2201.08239},
  
  author = {Thoppilan, Romal and De Freitas, Daniel and Hall, Jamie and Shazeer, Noam and Kulshreshtha, Apoorv and Cheng, Heng-Tze and Jin, Alicia and Bos, Taylor and Baker, Leslie and Du, Yu and Li, YaGuang and Lee, Hongrae and Zheng, Huaixiu Steven and Ghafouri, Amin and Menegali, Marcelo and Huang, Yanping and Krikun, Maxim and Lepikhin, Dmitry and Qin, James and Chen, Dehao and Xu, Yuanzhong and Chen, Zhifeng and Roberts, Adam and Bosma, Maarten and Zhao, Vincent and Zhou, Yanqi and Chang, Chung-Ching and Krivokon, Igor and Rusch, Will and Pickett, Marc and Srinivasan, Pranesh and Man, Laichee and Meier-Hellstern, Kathleen and Morris, Meredith Ringel and Doshi, Tulsee and Santos, Renelito Delos and Duke, Toju and Soraker, Johnny and Zevenbergen, Ben and Prabhakaran, Vinodkumar and Diaz, Mark and Hutchinson, Ben and Olson, Kristen and Molina, Alejandra and Hoffman-John, Erin and Lee, Josh and Aroyo, Lora and Rajakumar, Ravi and Butryna, Alena and Lamm, Matthew and Kuzmina, Viktoriya and Fenton, Joe and Cohen, Aaron and Bernstein, Rachel and Kurzweil, Ray and Aguera-Arcas, Blaise and Cui, Claire and Croak, Marian and Chi, Ed and Le, Quoc},
  
  title = {LaMDA: Language Models for Dialog Applications},
  
  publisher = {arXiv},
  
  year = {2022},
  
  copyright = {Creative Commons Attribution 4.0 International}
}

@article{gpt3_paper,
  author    = {Tom B. Brown and
               Benjamin Mann and
               Nick Ryder and
               Melanie Subbiah and
               Jared Kaplan and
               Prafulla Dhariwal and
               Arvind Neelakantan and
               Pranav Shyam and
               Girish Sastry and
               Amanda Askell and
               Sandhini Agarwal and
               Ariel Herbert{-}Voss and
               Gretchen Krueger and
               Tom Henighan and
               Rewon Child and
               Aditya Ramesh and
               Daniel M. Ziegler and
               Jeffrey Wu and
               Clemens Winter and
               Christopher Hesse and
               Mark Chen and
               Eric Sigler and
               Mateusz Litwin and
               Scott Gray and
               Benjamin Chess and
               Jack Clark and
               Christopher Berner and
               Sam McCandlish and
               Alec Radford and
               Ilya Sutskever and
               Dario Amodei},
  title     = {Language Models are Few-Shot Learners},
  journal   = {CoRR},
  volume    = {abs/2005.14165},
  year      = {2020},
  url       = {https://arxiv.org/abs/2005.14165},
  eprinttype = {arXiv},
  eprint    = {2005.14165},
  bibsource = {dblp computer science bibliography, https://dblp.org}
}

@misc{imagen,
  doi = {10.48550/ARXIV.2205.11487},
  
  url = {https://arxiv.org/abs/2205.11487},
  
  author = {Saharia, Chitwan and Chan, William and Saxena, Saurabh and Li, Lala and Whang, Jay and Denton, Emily and Ghasemipour, Seyed Kamyar Seyed and Ayan, Burcu Karagol and Mahdavi, S. Sara and Lopes, Rapha Gontijo and Salimans, Tim and Ho, Jonathan and Fleet, David J and Norouzi, Mohammad},
  
  title = {Photorealistic Text-to-Image Diffusion Models with Deep Language Understanding},
  
  publisher = {arXiv},
  
  year = {2022},
  
  copyright = {arXiv.org perpetual, non-exclusive license}
}

@misc{dalle2,
  doi = {10.48550/ARXIV.2204.06125},
  
  url = {https://arxiv.org/abs/2204.06125},
  
  author = {Ramesh, Aditya and Dhariwal, Prafulla and Nichol, Alex and Chu, Casey and Chen, Mark},
  
  title = {Hierarchical Text-Conditional Image Generation with CLIP Latents},
  
  publisher = {arXiv},
  
  year = {2022},
  
  copyright = {Creative Commons Attribution 4.0 International}
}

@article{foundation_models,
  author    = {Rishi Bommasani and
               Drew A. Hudson and
               Ehsan Adeli and
               Russ Altman and
               Simran Arora and
               Sydney von Arx and
               Michael S. Bernstein and
               Jeannette Bohg and
               Antoine Bosselut and
               Emma Brunskill and
               Erik Brynjolfsson and
               Shyamal Buch and
               Dallas Card and
               Rodrigo Castellon and
               Niladri S. Chatterji and
               Annie S. Chen and
               Kathleen Creel and
               Jared Quincy Davis and
               Dorottya Demszky and
               Chris Donahue and
               Moussa Doumbouya and
               Esin Durmus and
               Stefano Ermon and
               John Etchemendy and
               Kawin Ethayarajh and
               Li Fei{-}Fei and
               Chelsea Finn and
               Trevor Gale and
               Lauren Gillespie and
               Karan Goel and
               Noah D. Goodman and
               Shelby Grossman and
               Neel Guha and
               Tatsunori Hashimoto and
               Peter Henderson and
               John Hewitt and
               Daniel E. Ho and
               Jenny Hong and
               Kyle Hsu and
               Jing Huang and
               Thomas Icard and
               Saahil Jain and
               Dan Jurafsky and
               Pratyusha Kalluri and
               Siddharth Karamcheti and
               Geoff Keeling and
               Fereshte Khani and
               Omar Khattab and
               Pang Wei Koh and
               Mark S. Krass and
               Ranjay Krishna and
               Rohith Kuditipudi and
               et al.},
  title     = {On the Opportunities and Risks of Foundation Models},
  journal   = {CoRR},
  volume    = {abs/2108.07258},
  year      = {2021},
  url       = {https://arxiv.org/abs/2108.07258},
  eprinttype = {arXiv},
  eprint    = {2108.07258},
  bibsource = {dblp computer science bibliography, https://dblp.org}
}

@article{clark2021,
  author    = {Elizabeth Clark and
               Tal August and
               Sofia Serrano and
               Nikita Haduong and
               Suchin Gururangan and
               Noah A. Smith},
  title     = {All That's 'Human' Is Not Gold: Evaluating Human Evaluation of Generated
               Text},
  journal   = {CoRR},
  volume    = {abs/2107.00061},
  year      = {2021},
  url       = {https://arxiv.org/abs/2107.00061},
  eprinttype = {arXiv},
  eprint    = {2107.00061},
  bibsource = {dblp computer science bibliography, https://dblp.org}
}

@inproceedings{naacl_abbrev_paper,
  doi = {10.48550/ARXIV.2205.03767},
  
  url = {https://arxiv.org/abs/2205.03767},
  
  author = {Cai, Shanqing and Venugopalan, Subhashini and Tomanek, Katrin and Narayanan, Ajit and Morris, Meredith Ringel and Brenner, Michael P.},
  
  title = {Context-Aware Abbreviation Expansion Using Large Language Models},
  
  booktitle = {Proceedings of NAACL 2022},
  
  year = {2022},
  
  copyright = {Creative Commons Attribution 4.0 International}
}

@inproceedings{lampost, 
    url = {https://arxiv.org/pdf/2207.02308.pdf},
    
    author = {Goodman, Steven and et al.},
    
    title = {LaMPost: Design and Evaluation of an AI-assisted Email Writing Prototype for Adults with Dyslexia.},
    
    publisher = {ACM},
    
    year = {2022},
    
    booktitle = {Proceedings of ASSETS 2022}

}

@inproceedings{wordcraft,
author = {Yuan, Ann and Coenen, Andy and Reif, Emily and Ippolito, Daphne},
title = {Wordcraft: Story Writing With Large Language Models},
year = {2022},
isbn = {9781450391443},
publisher = {Association for Computing Machinery},
address = {New York, NY, USA},
url = {https://doi.org/10.1145/3490099.3511105},
doi = {10.1145/3490099.3511105},
booktitle = {27th International Conference on Intelligent User Interfaces},
pages = {841–852},
numpages = {12},
location = {Helsinki, Finland},
series = {IUI '22}
}

@misc{social_sim_uist,
  doi = {10.48550/ARXIV.2208.04024},
  
  url = {https://arxiv.org/abs/2208.04024},
  
  author = {Park, Joon Sung and Popowski, Lindsay and Cai, Carrie J. and Morris, Meredith Ringel and Liang, Percy and Bernstein, Michael S.},
  
  title = {Social Simulacra: Creating Populated Prototypes for Social Computing Systems},
  
  publisher = {arXiv},
  
  year = {2022},
  
  copyright = {Creative Commons Attribution 4.0 International}
}

@inproceedings{macromaker,
author = {Jiang, Ellen and Olson, Kristen and Toh, Edwin and Molina, Alejandra and Donsbach, Aaron and Terry, Michael and Cai, Carrie J},
title = {PromptMaker: Prompt-Based Prototyping with Large Language Models},
year = {2022},
isbn = {9781450391566},
publisher = {Association for Computing Machinery},
address = {New York, NY, USA},
url = {https://doi.org/10.1145/3491101.3503564},
doi = {10.1145/3491101.3503564},
booktitle = {Extended Abstracts of the 2022 CHI Conference on Human Factors in Computing Systems},
articleno = {35},
numpages = {8},
location = {New Orleans, LA, USA},
series = {CHI EA '22}
}

\end{document}